\documentclass[journal]{IEEEtran}
\pdfoutput=1 % ensures pdflatex on arxiv
\usepackage[utf8]{inputenc}
\usepackage{url}            % simple URL typesetting
\usepackage{booktabs}       % professional-quality tables
\usepackage{amsmath}       % blackboard math symbols
\usepackage{bbold}          % \mathbb{1}
\usepackage{nicefrac}       % compact symbols for 1/2, etc.
\usepackage{microtype}      % microtypography
\usepackage{nccmath}        % reduces size of equations
\usepackage{subfig}
\usepackage{enumitem}       % enumerate without large space
\usepackage{siunitx} % Required for alignment of numbers in the table. Use column type S instead of c.
\usepackage{tikz}
\usetikzlibrary{decorations.pathreplacing,calligraphy,positioning,calc,matrix,shapes,arrows,chains,shapes.geometric}
\usepackage[hidelinks,colorlinks = true,linkcolor = blue,
            urlcolor  = blue,
            citecolor = blue,
            anchorcolor = blue]{hyperref}       % hyperlinks 
\usepackage[big,compact]{titlesec} % Shrinks space around section headings
\usepackage[backend=bibtex,style=ieee,sorting=none,giveninits=true,maxbibnames=99,isbn=false,url=false,eprint=true,backref=true]{biblatex}

\renewbibmacro{in:}{}
\AtEveryBibitem{%
  \clearlist{language}%
}

\tikzset{
    between/.style args={#1 and #2}{
         at = ($(#1)!0.5!(#2)$)
    }
}

\title{Contrastive pretraining for semantic segmentation is \\robust to noisy positive pairs}

\author{Sebastian Gerard, Josephine Sullivan \thanks{This work is funded by
Digital Futures in the project EO-AI4GlobalChange.} \thanks{The authors work at KTH Royal Institute of Technology, 11428 Stockholm, Sweden. Mail: sgerard@kth.se, sullivan@kth.se} %\thanks{Mail: sgerard@kth.se, sullivan@kth.se} 
\thanks{This work has been submitted to the IEEE for possible publication. Copyright may be transferred without notice, after which this version may no longer be accessible.}
}

\begin{document}
%\iffalse
\maketitle

\begin{abstract}
    Domain-specific variants of contrastive learning can construct positive pairs from two distinct in-domain images, while traditional methods just augment the same image twice. For example, we can form a positive pair from two satellite images showing the same location at different times. Ideally, this teaches the model to ignore changes caused by seasons, weather conditions or image acquisition artifacts. However, unlike in traditional contrastive methods, this can result in undesired positive pairs, since we form them without human supervision. For example, a positive pair might consist of one image before a disaster and one after. This could teach the model to ignore the differences between intact and damaged buildings, which might be what we want to detect in the downstream task. Similar to false negative pairs, this could impede model performance. Crucially, in this setting only parts of the images differ in relevant ways, while other parts remain similar. Surprisingly, we find that downstream semantic segmentation is either robust to such badly matched pairs or even benefits from them. The experiments are conducted on the remote sensing dataset xBD, and a synthetic segmentation dataset for which we have full control over the pairing conditions. As a result, practitioners can use these domain-specific contrastive methods without having to filter their positive pairs beforehand, or might even be encouraged to purposefully include such pairs in their pretraining dataset.
\end{abstract}


\section{Introduction}

Contrastive learning can be used as self-supervised way to learn representations of images. For this, we define some image pairs as positive pairs. We teach the model to produce representations that are robust to the differences between those images, since they represent the same semantic content. All other pairs are defined as negative pairs, and their representations are encouraged to be sensitive to the difference between the respective images, since they differ in content. 

Non-specialized contrastive methods~\cite{chen_simple_2020, he_momentum_2020} construct positive pairs by sampling two differently augmented versions of the same image. Specialised methods use auxiliary \textit{domain-specific} data to create positive pairs from different in-domain images. For example, in remote sensing, various methods~\cite{leenstra_self-supervised_2021,manas_seasonal_2021} like Temporal Positives~\cite{ayush_geography-aware_2021} create positive pairs by sampling images from the same location taken at different times. This encourages representations to be robust towards irrelevant changes in seasons, weather conditions and image acquisition artifacts. Since we sample the positive pairs from locations in an unsupervised way, and the resulting data sets can be very big (e.g. 1.5TB for SSL4EO-S12~\cite{wang_ssl4eo-s12_2022}), some of these pairs are likely to contain changes that are relevant to the downstream task. For example, one image might be taken before, and one after a hurricane destroyed a building. These unknowingly included pairs could have a detrimental effect by encouraging representations to be robust to these relevant changes. We investigate whether this is the case and how big this effect is.

Previous work investigates the detrimental effect of false negative pairs~\cite{huynh_boosting_2022} or false positive pairs caused by random cropping~\cite{addepalli_towards_2021}(see \autoref{sec:related-work}) on downstream classification performance. The emergence of these domain-specific methods now creates the need to investigate the effect of false positive pairs in this setting. A crucial detail is that we investigate the effects on downstream segmentation, instead of classification, performance, due to the large relevance of segmentation tasks to change detection and earth observation in general. %This is crucial to the application of these methods, since e.g. floods or hurricanes strongly alter how a location appears on a satellite image. 

\begin{figure}[!t] % positioned here so it appears on the right column, not left before the abstract
    \centering
    \includegraphics[width=\columnwidth]{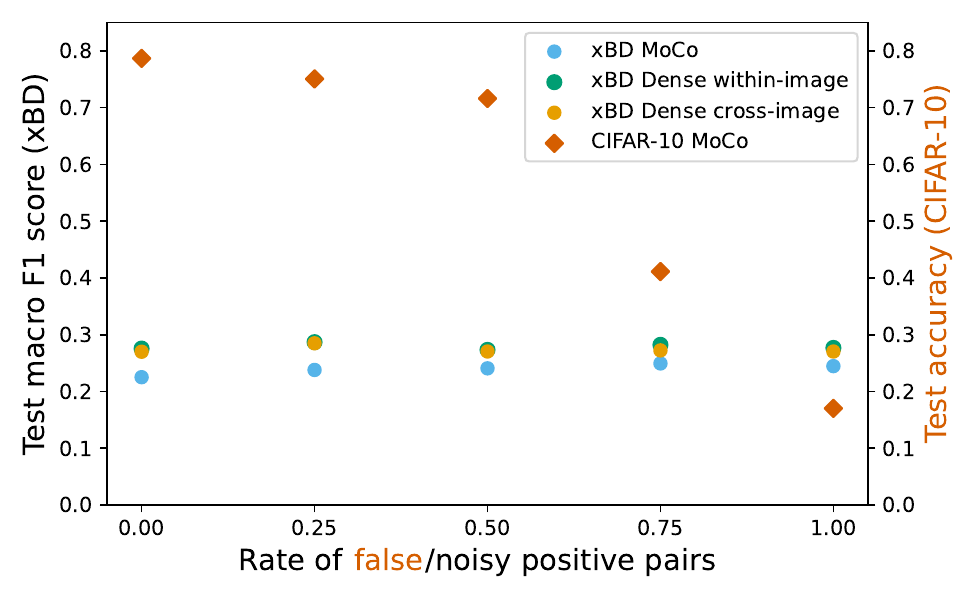}
    \caption{\textbf{False vs. noisy positives in contrastive learning} We mislabel CIFAR-10 images in contrastive pretraining, creating false positive pairs. These harm downstream classification. On the remote sensing dataset xBD, we pair images that only partly differ (e.g. buildings undamaged vs. damaged). This creates noisy pairs. Surprisingly, they do not harm downstream segmentation.} 
    \label{fig:1}
\end{figure}

During preliminary experiments on CIFAR10~\cite{krizhevsky_learning_2009} (see \autoref{fig:1}, details omitted), more false positive pairs in pretraining lead to worse downstream classification performance. When switching to a segmentation dataset in remote sensing  (xBD~\cite{gupta_creating_2019}) and a domain-specific creation of positive pairs (Temporal Positives~\cite{ayush_geography-aware_2021}), this does not hold true anymore. The key difference is that badly matched images in Temporal Positives are still partially correctly matched. After a hurricane, a house may be destroyed, while the lake next to it still looks the same as in the pre-hurricane image. That is why we refer to such image pairs as \textbf{noisy}, instead of false positive pairs.

We create a synthetic dataset (VTS), to more closely investigate how the characteristics of noisy pairs impact the downstream performance (see \autoref{sec:dataset}). We compare the results to those on xBD. We find that noisy positive pairs have a beneficial effect for the instance-based and dense contrastive losses we investigated (see \autoref{sec:contrastive-learning} for details). 
We furthermore investigate the source of the improvements in performance. We show that:
\begin{enumerate}[topsep=0pt,partopsep=0pt,itemsep=0pt]
    \item Contrastive pretraining on a real-world dataset for downstream semantic segmentation is robust to noisy positive pairs. See \autoref{sec:exp-xbd}.
    \item These findings can be confirmed on a synthetic dataset, on which we control the noise conditions. Here, the contrastive approaches (image level and dense) benefit from noisy pairs, except for extreme noise settings equivalent to false positive pairs. See \autoref{sec:exp-vts}.  
    \item This benefit is partly caused by an increased exposure to image features relevant to the downstream task. The performance improvements on the instance-based loss are additionally influenced by a strong regularization effect. See \autoref{sec:exp-xbd-exposure}.
\end{enumerate}

\section{Related work}
\label{sec:related-work}

\paragraph{False pairs are detrimental} The elimination of false negatives on ImageNet, when pretraining with SimCLR~\cite{chen_simple_2020}, was shown to lead to an improvement of 8.18\% in top-1 accuracy~\cite[Fig.10]{huynh_boosting_2022}. For CIFAR-10, the accuracy is improved by 4.5\%~\cite[Table 1]{addepalli_towards_2021}. Contrary to these works, we investigate noisy positive pairs, instead of false negative pairs. 

During BYOL~\cite{grill_bootstrap_2020} pretraining, random cropping can lead to false positive pairs~\cite{addepalli_towards_2021}. Filtering leads to an improvement of 3.5\%~\cite[Table 2]{addepalli_towards_2021}. This indicates that noisy positives might also be detrimental. However, in these false positive pairs, the whole crops are wrongly matched. In contrast, in our \textit{noisy} pairs, only parts of the images are falsely matched, while the rest of the matching is correct. 

\paragraph{Positive pairs made from distinct images} 
In remote sensing, the geospatial position of each image can be used to create positive pairs from two images taken at the same place, but different times~\cite{ayush_geography-aware_2021, leenstra_self-supervised_2021,manas_seasonal_2021}. 
In medical imaging, co-registered volumetric images of the same anatomic region in different patients can be used to create positive pairs~\cite{chaitanya_contrastive_2020}. In remote sensing a natural disaster happening between the acquisition of two images might lead to noisy positive pairs. In the same way, in medical imaging, two scans of different patients might be paired, showing the same region of the body, where one patient has an undiscovered malignant change. In this way, we believe our results to be relevant to both remote sensing and medical imaging.

A conceptually similar method exists for biodiversity monitoring via camera traps~\cite{pantazis_focus_2021}. Positive image pairs are created based on temporal or local proximity in multiple camera traps. Their performance only strongly drops when going beyond 75\% false positives. In contrast to this, our main focus lies on the effects of noisy (not false) positive pairs and the effect on segmentation (not classification) downstream performance.

\section{Contrastive learning}
\label{sec:contrastive-learning}

\begin{figure}
\centering
%\resizebox{\columnwidth}{!}{

\textbf{1. Contrastive pretraining \vspace*{0.2cm}\\}
\begin{tikzpicture}[thick]

    \node[inner sep=0pt] (pos_image1) at (0,0) 
    {\includegraphics[width=0.2\columnwidth]{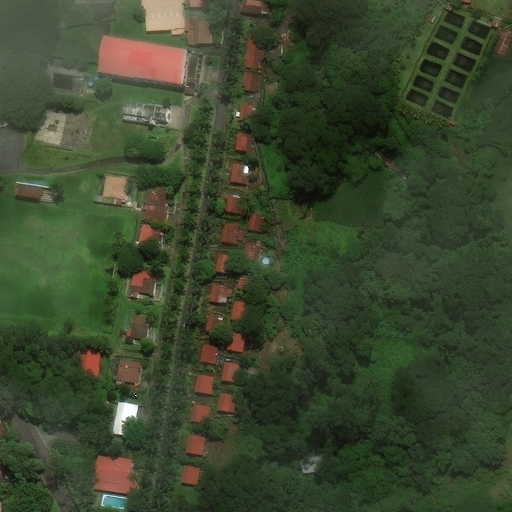}};
    \node[inner sep=0pt] (pos_image2) at (2.5,0) 
    {\includegraphics[width=0.2\columnwidth]{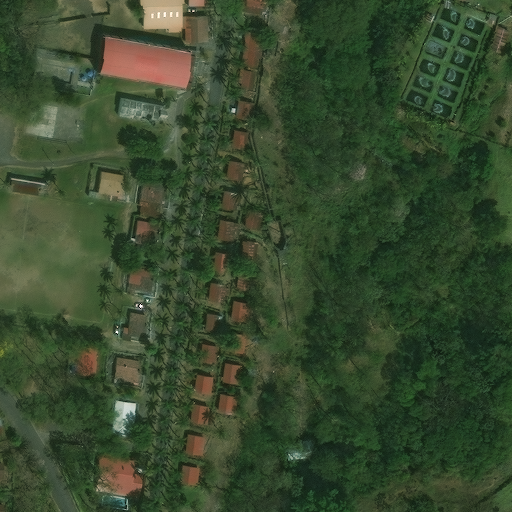}};
    \node[inner sep=0pt] (neg_image) at (5,0) 
    {\includegraphics[width=0.2\columnwidth]{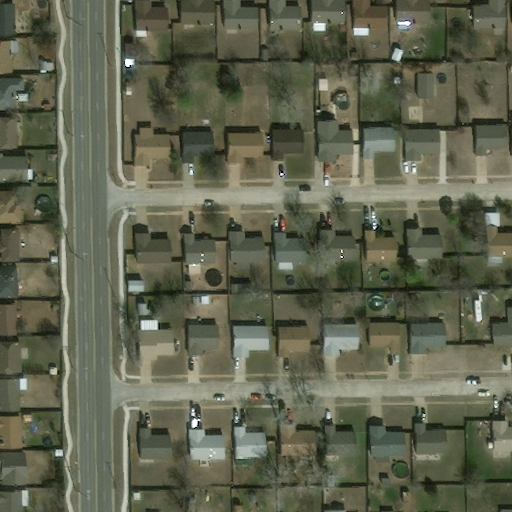}};

    \node[above = 0.0cm of pos_image1] (aboveimage1) {$\hat{I}$};
    \node[above = 0.0cm of pos_image2] (aboveimage2) {$I$};
    \node[above = 0.0cm of neg_image] (aboveimage3) {$J$};

    \draw [decorate, decoration = {brace}] (aboveimage1) --  (aboveimage2);

    \draw [decorate, decoration = {brace}] (aboveimage2) --  (aboveimage3);
    
    \node[above = 0.0cm of $(aboveimage1.north)!0.5!(aboveimage2.north)$] (betweenimage12) {positive pair};
    \node[above = 0.0cm of $(aboveimage2.north)!0.5!(aboveimage3.north)$] (betweenimage23) {negative pair};

    %\node[below=0.5cm of pos_image1, fill=green!30] (encoder1) {encoder $\theta$};
    \node[below=0.5cm of pos_image1, draw, fill=green!30, trapezium, trapezium angle=-70, inner xsep=3pt, inner ysep=10pt] (encoder1) {encoder $\theta$};
    \node[below=0.5cm of pos_image2, draw, fill=green!30, trapezium, trapezium angle=-70, inner xsep=3pt, inner ysep=10pt] (encoder2) {encoder $\theta$};
    \node[below=0.5cm of neg_image, draw, fill=green!30, trapezium, trapezium angle=-70, inner xsep=3pt, inner ysep=10pt] (encoder3) {encoder $\theta$};

    \draw[-latex'] (pos_image1.south) -- (encoder1.north);
    \draw[-latex'] (pos_image2.south) -- (encoder2.north);
    \draw[-latex'] (neg_image.south) -- (encoder3.north);

    \node[below=0.5cm of encoder1] (f1) {$f(\hat{I})$};
    \node[below=0.5cm of encoder2] (f2) {$f(I)$};
    \node[below=0.5cm of encoder3] (f3) {$f(J)$};

   \draw[-latex'] (encoder1.south) -- (f1.north);
   \draw[-latex'] (encoder2.south) -- (f2.north);
   \draw[-latex'] (encoder3.south) -- (f3.north);

    \node[below=0.5cm of f1.south] (below_f1) {};
    \node[] (below_f2) at (below_f1 -| f2) {};
    \node[] (below_f3) at (below_f2 -| f3) {};
    
    \node[between = below_f1 and below_f2, fill=blue!30, align=center, draw] (loss12) {contrastive\\loss};
    \node[between = below_f2 and below_f3, fill=red!30, align=center, draw] (loss23) {contrastive\\loss};

    \draw[-latex'] (f1.south) |- (loss12);
    \draw[-latex'] (f2.south) |- (loss12);
    \draw[-latex'] (f2.south) |- (loss23);
    \draw[-latex'] (f3.south) |- (loss23);

    \node[below=0.0cm of loss12] (a) {(attract)};
    \node[below=0.0cm of loss23] (b) {(repel)};
    
\end{tikzpicture}%}
\textbf{\\\vspace*{0.3cm}2. Train task-specific segmentation head \\based on pre-trained encoder\\\vspace*{0.3cm}}
\begin{tikzpicture}[thick]

    \vspace{1cm}
    %\node[] at (0,-1){\,};
    \node[inner sep=0pt] (image) at (0,0) 
    {\includegraphics[width=0.2\columnwidth]{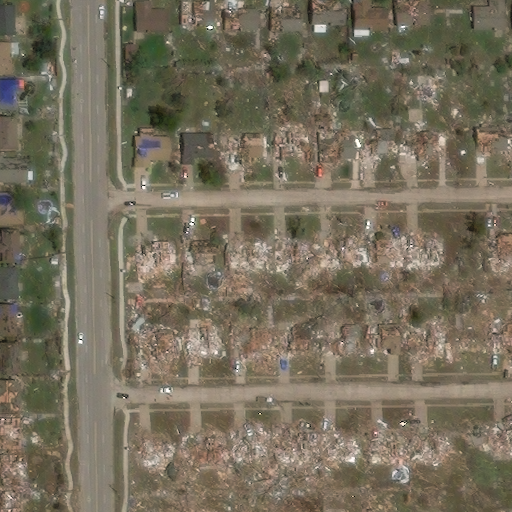}};

    \node[align=center, right =2.4cm of image, fill=green!30, draw, trapezium, rotate=90, trapezium angle=-70,anchor=south, inner xsep=4pt, inner ysep=4pt] (encoder) {\rotatebox[]{-90}{encoder $\theta$}};
    \node[below=0.3 of encoder.west] (frozen_encoder) {(frozen)};
    \node[align=center, right = 4.0cm of image, draw, fill=black!25, trapezium, rotate=90, trapezium angle=70,anchor=south, inner xsep=4pt, inner ysep=4pt] (segmentation_head) {\rotatebox[]{-90}{head}};
    \node[inner sep=0pt, right = 4.7cm of image] (label) 
    {\includegraphics[width=0.2\columnwidth]{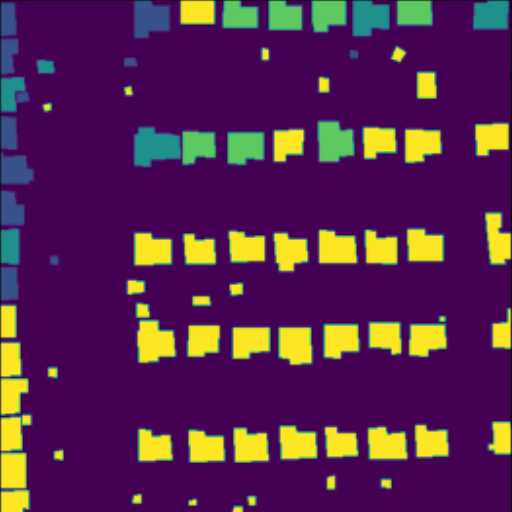}};

    \draw[-latex'] (image) -- (encoder);
    \draw[-latex'] (encoder) -- (segmentation_head);
    \draw[-latex'] (segmentation_head) -- (label);
    
\end{tikzpicture}
\\\vspace{0.3cm}
\caption{\textbf{Contrastive learning} learns by comparing feature representations $f(\cdot)$. Pairs of representations defined as similar (positive pairs) are pushed closer together by the contrastive loss. Those defined as dissimilar (negative pairs) are pushed apart. For clarity, this illustration omits the details regarding the projection head, momentum encoder and queue of negatives that are used in MoCo.
After contrastive pretraining, we freeze the encoder and train a segmentation head on top of it in a supervised way. This allows us to evaluate the quality of the learned representations. In an application scenario, we would finetune the whole network, including the encoder, instead.} 
\label{fig:contrastive-basics}
\end{figure}
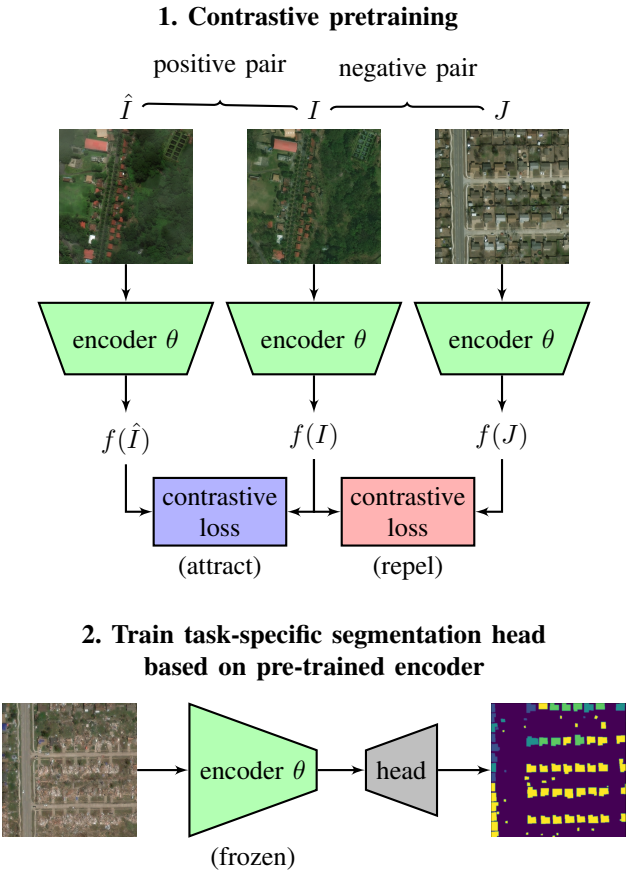

The basic idea of contrastive learning is to learn by comparing feature representations. We define some pairs of representations as positive, other pairs as negative. The loss function encourages positive pairs to move closer together and negative pairs to move further apart. After pretraining a model using contrastive learning with an unlabeled dataset, we finetune the model on a labeled dataset. In practice, this labeled dataset can be much smaller than the unlabeled dataset, thereby saving a lot of work that would be required for labeling. We instead use the full labeled dataset in finetuning, to ensure that we measure the quality of the learned representations instead of effects caused by the amount of data used in finetuning. We freeze the pretrained model and only train a segmentation head on top, to evaluate the quality of the learned representations. Again, this is to make sure that we evaluate the learned representations. In practical applications, the whole model, including the encoder, would be finetuned. \autoref{fig:contrastive-basics} visualizes these basic concepts.

We use two different contrastive algorithms in our experiments. First is the widely-used MoCo v2~\cite{chen_improved_2020} algorithm. It is based on comparing feature vectors describing whole images.
The images we pair show the same location at two different times, which is the Temporal Positives~\cite{ayush_geography-aware_2021} setup. Secondly, we use two variants of a dense contrastive loss, described below. The dense losses compare feature vectors describing sub-regions of images, instead of whole images.

\subsection{Instance-discrimination: Comparisons at the image level}

Recent works applying self-supervised learning for pretraining computer vision models tend to follow the paradigm of \emph{instance discrimination}~\cite{wu_unsupervised_2018}. In this paradigm, each individual image (instance) is treated as its own class. The learning task is to discriminate between images representing the same class and images representing different classes. Positive pairs therefore consist of augmented versions of the same image, while negative pairs consist of augmented versions of different images. We use the widely-used MoCo v2~\cite{chen_improved_2020} framework to pretrain neural networks.

The corresponding loss is called InfoNCE~\cite{oord_representation_2019}. We adapt the notation in  \cite{he_momentum_2020} slightly, to define it as: 

\begin{equation}
\label{eq:infoNCE}
%\medmath{
%\mathcal{L}_{\mathrm{InfoNCE}}=
%    -\log \frac{\exp \left(q \cdot k_{+} / \tau\right)}{\sum_{i=0}^{K} \exp \left(q \cdot k_{i} / \tau\right)}
    -\log \frac{\exp \left(f(I) \cdot f(\hat{I}) / \tau\right)}{\sum_{i=0}^{Q} \exp \left(f(I) \cdot q_{i} / \tau\right)}
 %   }
\end{equation}

Here, $I$ and $\hat{I}$ are two images belonging to a positive pair. Conventionally, $I$ and $\hat{I}$ are generated from a single source image by applying two different random augmentations to it. We instead pair two satellite images showing the same location at different times, which could be interpreted as a natural form of augmentation. $f(I)$ is the feature representation of image $I$. $\tau$ is the softmax temperature hyperparameter. In addition to the images in a batch, MoCo v2 keeps a queue of feature representations $\{q_{i}| 0\leq i\leq Q\}$  of previous images. They form negative pairs with image $I$, since they represent other images. For further details, please refer to the original papers~\cite{he_momentum_2020,chen_improved_2020}.

\subsection{Dense contrastive learning: Comparisons at the patch-level}
\label{sec:method-dense}

MoCo uses feature representations of \textit{full images}. In this section, we contrast feature representations of \textit{sub-regions} within an image. Such methods are sometimes referred to as \textbf{dense contrastive} methods~\cite{wang_dense_2021, zhang_looking_2021}. The goal is to force the network to pay attention to local features, instead of global ones. This is useful~\cite{zhao_contrastive_2021} when contrastive learning is used as pretraining for semantic segmentation tasks, which require more fine-grained features. 

We use the supervised dense contrastive method presented in \cite{zhao_contrastive_2021}, which we describe below. While we are generally interested in self-supervised approaches, this dense method uses full segmentation labels. We can use the results as an upper bound for how well dense contrastive methods (including self-supervised ones) can cope with noisy positive pairs.

For each input image $I$, let $\hat{I}$ be the image that it forms a Temporal Positive image pair with. We extract both $d{\times}d$ feature maps $f(I), f(\hat{I})$ and down-sample both images' ground truth segmentation labels to $d{\times}d$. Using the down-sampled labels $y^I, y^{\hat{I}}$, we can create positive and negative pairs between each of the $d^2$ pixel positions. Positive pairs are formed between pixel positions assigned to the same class, negative pairs are formed between pixel positions belonging to different classes. 

$p \in [1, N^I]$ indexes the $N^I = d^2$ pixel-wise features of $I$'s feature map $f(I)$, $q$ indexes $f(\hat{I})$. 
Each of those features is a unit-normalized vector. $N_{y_{p}^{I}}^{\hat{I}}$ is the number of pixels in $f(\hat{I})$ with class label $y^I_p$. The indicator function $\mathbb{1}_{p q}^{I \hat{I}}$ is 1 if 
$y^I_p = y^{\hat{I}}_q$, and 0 otherwise. We use a short-hand to write $\exp(f(I)_p \cdot f(\hat{I})_q / \tau) = e_{pq}^{I\hat{I}}.$  

We use two dense loss functions. The first is called \textbf{within-image loss}~\cite[Eq. 1]{zhang_deep_2021}. Pairs are created between two augmented versions of the same image. 

\begin{equation}
\label{eq:within_img}
%\medmath{
    %\mathcal{L}_{\mathrm{within}} = 
    -\frac{1}{N^{I}} \sum_{p=1}^{N^{I}} \frac{1}{N_{y_{p}^{I}}^{\hat{I}}} \sum_{q=1}^{N^{\hat{I}}} \mathbb{1}_{p q}^{I \hat{I}} \log \left(\frac{e_{p q}^{I \hat{I}}}{\sum_{k=1}^{N^{\hat{I}}} e_{p k}^{I \hat{I}}}\right)
%}
\end{equation}

The second loss function is called \textbf{cross-image loss}~\cite[Eq. 2]{zhang_deep_2021}. It extends the within-image loss by a component that pairs pixels in $f(I)$ with pixels in $f(\hat{J})$, where I and J are different images and do not form a Temporal Positive pair. %On the implementation level, we take a mini-batch of images and their assigned pair-partner, and then shuffle which partner-image $\hat{J}$ is matched to which image $I$ for the cross-image pairs.

\begin{equation}
%\medmath{
\begin{aligned}
%&\mathcal{L}_{\mathrm{cross}} = \\
&-\frac{1}{N^{I}} \sum_{p=1}^{N^{I}} \sum_{q=1}^{N^{\hat{I}}} \frac{\mathbb{1}_{p q}^{I \hat{I}}}{N_{y_{p}^{I}}^{\hat{I}}+N_{y_{p}^{I}}^{\hat{J}}}
\log \left(\frac{e_{p q}^{I \hat{I}}}{\sum_{k=1}^{N^{\hat{I}}} e_{p k}^{I \hat{I}}+\sum_{k=1}^{N^{\hat{J}}} \mathbb{1}_{p k}^{I \hat{J}} e_{p k}^{I \hat{J}}}\right)\\
&-\frac{1}{N^{I}} \sum_{p=1}^{N^{I}} \sum_{q=1}^{N^{\hat{J}}} \frac{\mathbb{1}_{p q}^{I \hat{J}}}{N_{y_{p}^{I}}^{\hat{I}}+N_{y_{p}^{I}}^{\hat{J}}}
\log \left(\frac{e_{p q}^{I \hat{J}}}{\sum_{k=1}^{N^{\hat{I}}} e_{p k}^{I \hat{I}}+\sum_{k=1}^{N^{\hat{J}}} \mathbb{1}_{p k}^{I \hat{J}} e_{p k}^{I \hat{J}}}\right)
\end{aligned}%}
\end{equation}

\section{Datasets}
\label{sec:dataset}
We use two different datasets to investigate the effect of noisy positive pairs, xBD and VTS, which is a synthetic dataset to investigate the influence of noise conditions in noisy positive pairs more thoroughly.

For xBD, we can include or exclude noisy positive pairs in our pretraining. We call this varying the \textbf{noisy pairs rate} $r_{pairs} \in [0, 1]$. For example, $r_{pairs} {=} 0.7$ means that 70\% of all positive pairs in pretraining are noisy pairs. For VTS, we additionally control how many pixels per image pair are wrongly matched. We call this the \textbf{per-image noise} $r_{img} \in [0, 1]$. For example, $r_{img} {=} 0.7$ means that, on average, 70\% of all pixel positions in a noisy image pair do not contain the same semantic information in both images, but differ in a way that is relevant to the downstream task.

\subsection{xBD}
\label{sec:dataset-xBD}

The xBD dataset~\cite{gupta_creating_2019} is a remote sensing dataset, consisting of high-resolution, bi-temporal satellite images and high-quality expert labels. The task it represents is the gradual classification of building damage after natural events like wildfires, hurricanes or floods, using a pre- and a post-event image. An example image pair can be seen in \autoref{fig:exposure-explained}.%Each pixel is annotated as belonging to one of five classes: one background class, one class representing undamaged buildings and three classes representing damaged buildings of varying severity.

We cut each $1024{\times}1024$ pixels image into four $512{\times}512$ pixels images without overlap. We use xBD's \texttt{train} and \texttt{tier3} subsets. We create a 70/30-split into train/val sets, stratifying over disaster sites. 
This yields 20\,446 clean and 5\,224 noisy pairs in the training set (see below). They are under-sampled to achieve the desired proportions of noisy and noise-free pairs in pretraining. During finetuning, all images are used. Class-weighting is used to account for data imbalance. The validation set always contains the same 11\,002 image pairs. The test set consists of 3\,732 images, which is the official test set, cut up into $512{\times}512$ pixel images. We use augmentations from the winning solution~\cite{durnov_xview2_2020} of the xView2 challenge, which is based on xBD.

\subsubsection{Noisy positive pairs}
\label{sec:xbd-noisy}
For xBD, a noisy pair consists of a pre- and a post-disaster image from the same location, where the post-disaster image contains damaged buildings, but the pre-disaster image does not. These positive pairs encourage the model to output the same representation for both images. The model might learn to ignore differences between damaged and undamaged buildings. However, the downstream task requires us to distinguish between these. Therefore, we call such positive pairs noisy. 

When varying the noisy pairs rate $r_{pairs}$, we under-sample clean or noisy positive pairs, until the desired ratio is reached. For example, for $r_{pairs} {=} 0.1$, we use all the clean pairs and add noisy pairs until these make up 10\% of the new dataset. 

For the dense losses, we use labels to match image patches with the same label. We binarize the five-class labels into background vs building pixels. This mimics a realistic setting in which we have rough localization information and creates a noisy matching, similar to Temporal Positives with MoCo.

\subsection{Synthetic Voronoi-cell texture segmentation dataset (VTS)}
\label{sec:dataset-voronoi}

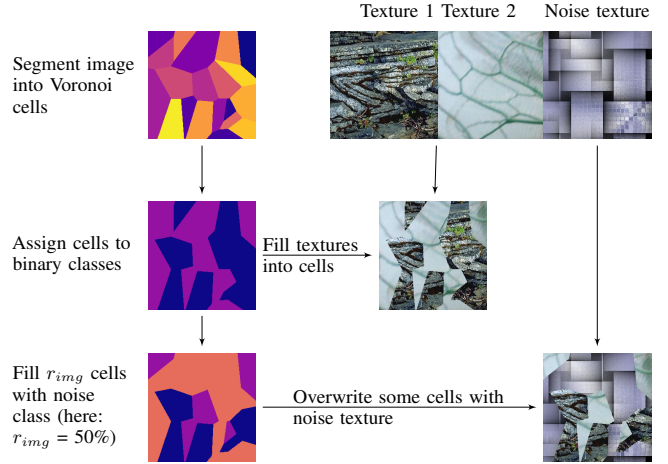
\begin{figure}[!t]
    \centering
    \resizebox{\columnwidth}{!}{
    \begin{tikzpicture}
        \small
        \node[] (vcells) at (0,0) {\includegraphics[width=0.2\columnwidth]{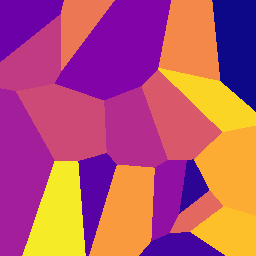}};
        \node[] (vcells_binary) at (0,-2.8) {\includegraphics[width=0.2\columnwidth]{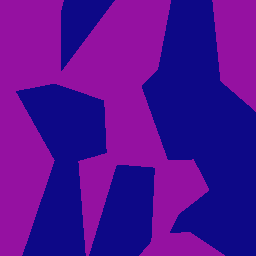}};
        \node[] (vcells_noisy) at (0,-5.3) {\includegraphics[width=0.2\columnwidth]{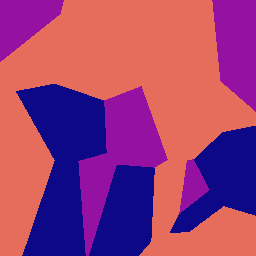}};
        
        \node[right= 1cm of vcells] (textures12)  {\includegraphics[width=0.2\columnwidth]{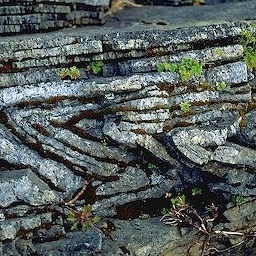}%
        \includegraphics[width=0.2\columnwidth]{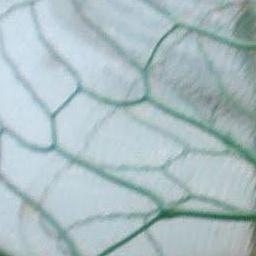}};
        \node[right= 4.5cm of vcells] (texture3) {\includegraphics[width=0.2\columnwidth]{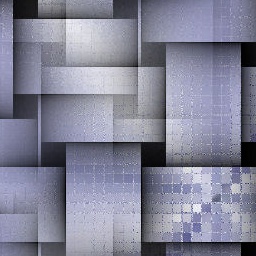}};
        
        \node[right= 1.8cm of vcells_binary] (composed_binary) {\includegraphics[width=0.2\columnwidth]{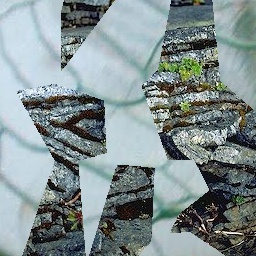}};
        
        \node[right= 4.5cm of vcells_noisy] (composed_noisy) {\includegraphics[width=0.2\columnwidth]{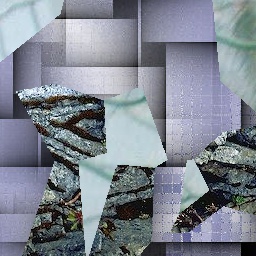}};
        
        \path [draw, -latex'] (vcells_binary) -- node [text width=1.8cm,midway] {Fill textures
into cells} (composed_binary);
        \path [draw, -latex'] (vcells_noisy) -- node [text width=3.5cm,midway] {Overwrite some cells
        with noise texture} (composed_noisy);
        \draw[-latex'] (texture3) -- (composed_noisy);
        \draw[-latex'] (textures12) -- (composed_binary);
        
        \draw[-latex'] (vcells) -- (vcells_binary);
        \draw[-latex'] (vcells_binary) -- (vcells_noisy);

        \node[above= 0.0cm of textures12]{Texture 1  Texture 2};
        \node[above= 0.0cm of texture3]{Noise texture};
        
        \node[left= 0.0cm of vcells,text width=2cm]{Segment image into Voronoi cells};
        \node[left= 0.0cm of vcells_binary,text width=2cm]{Assign cells to binary classes};
        \node[left= 0.0cm of vcells_noisy,text width=2cm]{Fill $r_{img}$ cells with noise class (here: $r_{img}$ = 50\%)};
    \end{tikzpicture}
    }
    \caption{\textbf{Generation process of the VTS dataset:} The $256{\times}256$ image is segmented into 20 random Voronoi cells, which are evenly randomly distributed between two classes. The cells are filled with a texture images representing the respective classes. A noisy version of the image is created by uniformly at random filling $r_{img}$ of the cells with a noise texture. Noisy positive pairs are created by pairing the noise-free image consisting of only two textures with the noisy image, consisting of three textures. The noise texture can be seen analogously to the damaged buildings in the xBD dataset.}
    \label{fig:voronoi-generation}
\end{figure}

We generate a synthetic segmentation dataset, which allows for full control over the conditions of noisy positive pairings. \autoref{fig:voronoi-generation} visualizes the dataset generation process. We randomly segment each to-be-generated image into 20 Voronoi cells, which we equally separate randomly between two classes. The two classes are represented by two different texture categories in the Describable Textures Dataset (DTD)~\cite{cimpoi_describing_2014}. Each texture category contains 120 images.  We use the non-overlapping categories \textit{stratified} and \textit{veined}. For each image that we generate, we pick one texture image from each of the two texture categories. Since the images differ in size, we first randomly crop out an area of the same size as the image we generate. We now fill all cells belonging to the first class, by copying the pixels from the texture image of the first texture category into the cells assigned to this class. We repeat the process for the second texture category, to fill the remaining cells.

The downstream task for VTS is a three-class semantic segmentation, including a noise texture class (see \autoref{sec:vts-noisy}). This is akin to xBD, where the damaged houses also need to be recognised in the downstream segmentation task. The noise class corresponds to the damaged houses. To create a train/validation/test split, we split the texture images at ratios of 0.5/0.3/0.2. We generate 6\,000/3\,600/2\,400 images. We use image augmentations similar to those used for xBD.

\subsubsection{Noisy positive image pairs}
\label{sec:vts-noisy}

To create noisy pairs, we pair a noise-free image with an image in which $r_{img}$ of Voronoi cells are filled with a noise texture (texture class \textit{matted}). Cells to replace are chosen uniformly at random, without regard for class balance. During training, each time we load an image, we pair it with its noisy partner with a probability given as the noisy pairs rate $r_{pairs}$, and we pair it with a differently augmented version of itself with probability $1-r_{pairs}$.

For the dense supervised losses, we use the noise-free label for both the noise-free and the noisy images. This is similar to the xBD dataset, where we assume that both images show the same information, without any differences in label information, even though some buildings might be (unbeknownst to the practitioner) damaged.

\section{Experiments}
\label{sec:experiments}
For all experiments, we first pretrain the encoder model with a contrastive loss (MoCo, within-image or cross-image). The noisy pairs used in pretraining are characterized by $r_{pairs}$ and $r_{img}$, which we vary between experiments, to detect how these parameters relate to the downstream performance. We keep the model weights with the best validation loss in pretraining. The validation set is always free of noise. We add a segmentation head after the final convolutional layer of the frozen pretrained encoder. The segmentation head is trained on the noise-free training set and evaluated on the test set. This training procedure is also visualized in \autoref{fig:contrastive-basics}.

\subsection{Implementation details}

Our goal is to single out the potential negative influence of noisy positive pairs. Therefore, we purposefully limit the model's capacity, using a simple ResNet18 as our encoder architecture with an FCN~\cite{shelhamer_fully_2017} segmentation head. For dense losses, we extract the $d{\times}d$ feature map $f(I)$ right before the global average pooling operation. In our experiments, $d$ is 8 for VTS and 16 for xBD. 

For xBD, we only provide the post-disaster image to the segmentation head. This forces the network to rely on its learnt features, instead of exploiting differences between pre- and post-disaster image. These choices naturally lead to worse performance. However, we are only interested in detecting \textit{relative} performance changes caused by noisy pairs, not in achieving the best absolute performance. During finetuning, we use class weights equal to the inverse class frequency. All numbers reported are based on single runs. Preliminary experiments indicate that the test performance among different conditions of randomness usually spans less than 1\%. Hyperparameters were largely taken from \cite{ayush_geography-aware_2021} for xBD and MoCo, and from \cite{zhao_contrastive_2021} for dense losses. The learning rates were determined via a coarse grid search.

\subsection{Contrastive pretraining for segmentation is robust to noisy pairs}
\label{sec:exp-xbd}
We vary the amount of noisy positive pairs in pretraining on xBD and measure the resulting downstream segmentation performance. The results are visualized in \autoref{fig:1}. MoCo benefits slightly from the inclusion of noisy pairs (up to 2.4\%), while the dense losses can benefit up to 1.4\%, but also lose up to 0.2\%. Overall, the downstream performance is not very sensitive to varying the amount of noisy pairs. This is especially true when comparing our results with those on CIFAR-10 (see \autoref{fig:1}) that are very sensitive to the inclusion of false positive pairs.

A simple first explanation for this robustness would be that the pretraining already does not work well, so the noise can not have much of an influence. However, while our best test F1 score is 0.287, ImageNet-pretraining achieves only 0.254, verifying our pretraining's suitability.

\phantomsection\label{sec:exp-vts}
The result that noisy pairs do not degrade downstream performance on xBD is surprising. An explanation could be that the noisy matches (damaged vs. undamaged buildings) take up too little space in the image. They would therefore have little influence on the loss function, too. Since we can not appropriately change this on xBD, we create the VTS dataset (see \autoref{sec:dataset-voronoi}). It allows us to control how many pixels per image pair are wrongly matched ($r_{img}$, analogous to wrongly matched building pixels), in addition to how many positive pairs are impacted by noisy matches ($r_{pairs}$).

\begin{figure}
    \centering
    \includegraphics[width=\columnwidth]{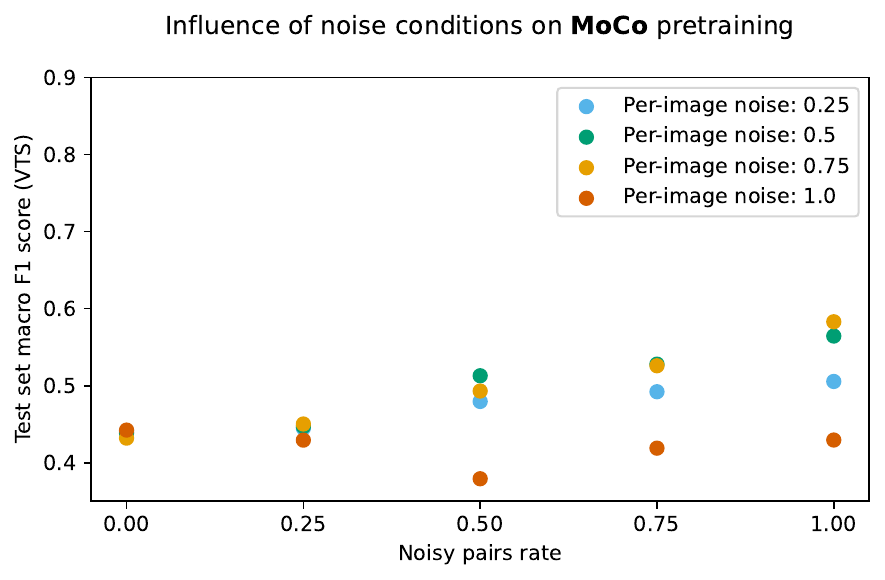}
    \caption{\textbf{VTS: MoCo benefits from the inclusion of noisy positive pairs in contrastive pretraining.} More noisy pairs lead to greater improvements, except for per-image noise $r_{img}{=}1.0$, in which completely different images are paired with each other.}
    \label{fig:exp-voronoi-moco}
\end{figure}

\begin{figure}
    \centering
    \includegraphics[width=\columnwidth]{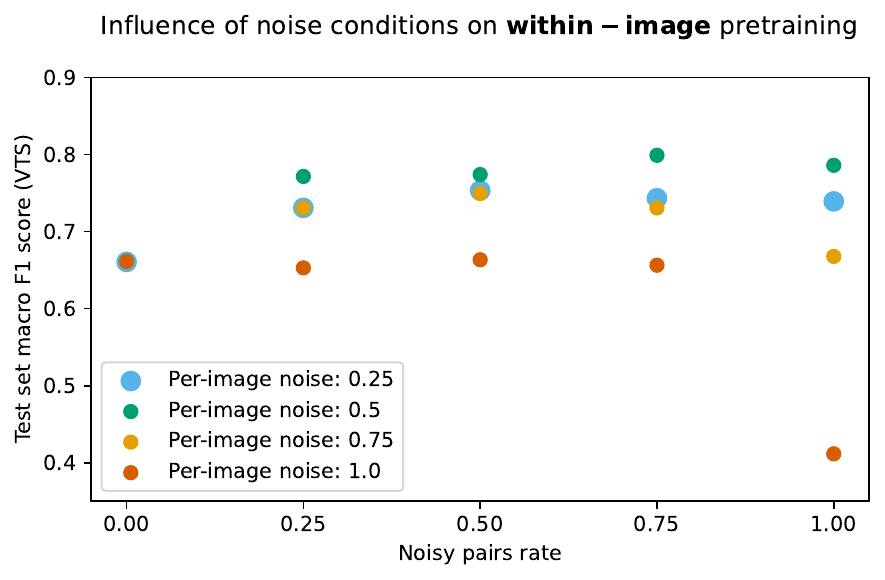}
    \includegraphics[width=\columnwidth]{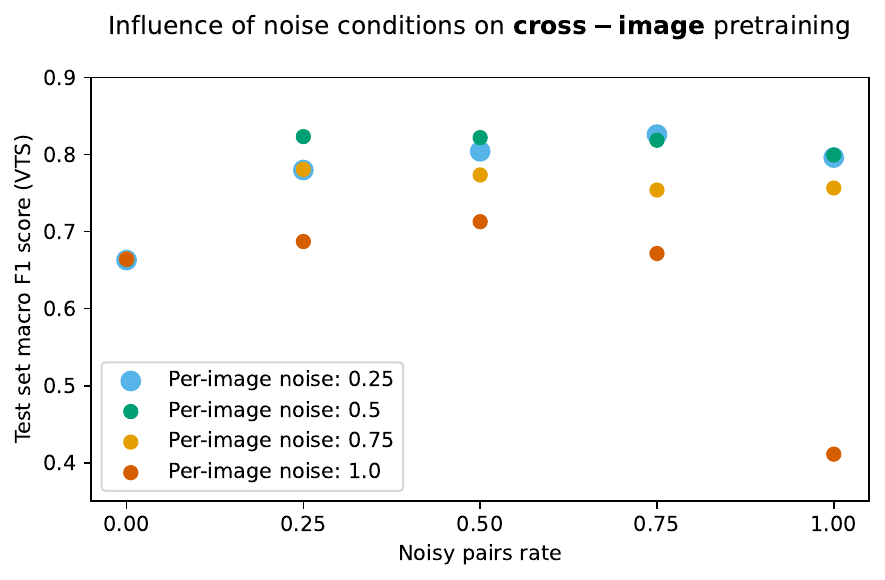}
    \caption{\textbf{VTS: Dense losses benefit from the inclusion of noisy positive pairs in contrastive pretraining.} More noisy pairs usually do not lead to greater improvements. Per-image noise of 1.0 can lead to a decline in quality, which makes intuitive sense, since it represents pairing two completely different images with each other. Points belonging to \textit{Per-image noise: 0.25} are enlarged for better visibility.}
    \label{fig:exp-voronoi-dense}
\end{figure}

\paragraph{VTS: Noisy pairs increase performance.}
We repeat the experiments on VTS, varying both $r_{pairs}$ and $r_{img}$. The results for MoCo can be seen in \autoref{fig:exp-voronoi-moco}. The results for the dense losses can be seen in \autoref{fig:exp-voronoi-dense}.
MoCo's performance increases with rising $r_{pairs}$. Dense losses also benefit from noise, but increasing $r_{pairs}$ above 25\% does not yield additional gains in most experiments. The relation between $r_{img}$ and the downstream performance is less clear. Only the extreme case of $r_{img}{=}100\%$, which is equivalent to false positives, leads to no improvement or even a degradation, with large drops for dense losses if $r_{img}{=}r_{pairs}{=}100\%$.
These results align with those on xBD. Noisy pairs are usually helpful, not harmful. In the next section, we explore what the cause of this might be.

\subsection{Regularization or exposure?}
\label{sec:reguarlization}

\begin{figure}
\centering
\resizebox{\columnwidth}{!}{
\begin{tikzpicture}[thick]
    \node[inner sep=0pt] (image1) at (0,0) 
    {\includegraphics[width=0.22\columnwidth]{images/moore-tornado_00000110_pre_disaster_1_0.png}};
    \node[inner sep=0pt] (image2) at (3.25,0) 
    {\includegraphics[width=0.22\columnwidth]{images/moore-tornado_00000110_post_disaster_1_0.png}};
    \node[inner sep=0pt] (image3) at (6.5,0) 
    {\includegraphics[width=0.22\columnwidth]{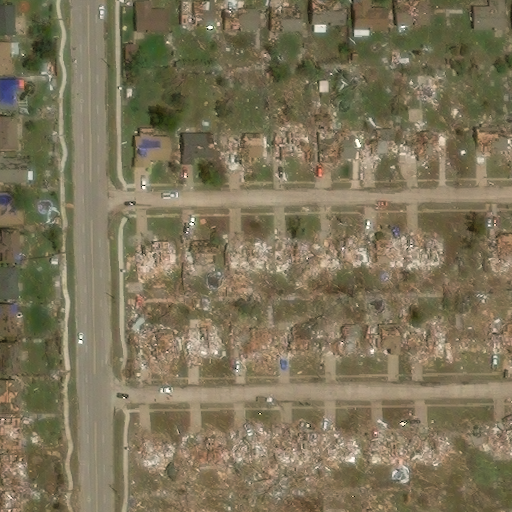}};
    
    \node[below = 0.0cm of image1] (underimage1) {};
    \node[below = 0.0cm of image2] (underimage2) {};
    \node[below = 0.0cm of image3] (underimage3) {};
    
    \node[below = 0.0cm of image1] (aboveimage1) {};
    \node[below = 0.0cm of image2] (aboveimage2) {};
    \node[below = 0.0cm of image3] (aboveimage3) {};
    
    \draw (image1) edge[<->] node [text width=1.05cm,midway,align=center] {\footnotesize Noisy positive} (image2);
    \draw (image2) edge[<->] node [text width=1.05cm,midway,align=center] {\footnotesize Mere exposure} (image3);

    \node[between = image1 and image2] (betweenimage12) {};
    \node[between = image2 and image3] (betweenimage23) {};
    
    \node[above = 0.0cm of image1,text depth=0pt] {\footnotesize Pre-disaster};
    \node[above = 0.0cm of image2,text depth=0pt] {\footnotesize Post-disaster};
    \node[above = 0.0cm of image3,text depth=0pt] {\footnotesize Post-disaster};
\end{tikzpicture}}
\caption{\textbf{Noisy positives vs. mere-exposure} Noisy positives pair a pre- with a post-disaster image. Mere-exposure pairs use the same post-disaster image twice. All images are independently augmented. Images are from the xBD dataset.\protect\footnotemark} %TODO check whether footnote and figure are still on same page before submission 
\label{fig:exposure-explained}
\end{figure}

The downstream task on xBD is to discern damaged from undamaged buildings. However, when we remove all noisy pairs from pretraining, the network needs to recognise damaged buildings in the downstream task without ever having seen any while pretraining the encoder. Re-introducing noisy pairs exposes the model to these crucial features and should enable it to better recognise building damage. We investigate whether the gains on xBD and VTS are merely the effect of being exposed to these relevant features or whether there is an additional effect at play. For this, we replace the noisy pairs, constructed with a pre- and a post-disaster image, with pairs consisting of two augmented versions of the same post-disaster image (see \autoref{fig:exposure-explained}). This exposes the network to the relevant features without noisy learning signals. We call this the \textbf{mere-exposure} setting. It retains the exposure element, while removing the aspect of noise in pairing that the noisy positives have introduced. We compute the performance difference between pretraining on noisy positives and mere exposure pairs.

\begin{figure}[!t]
    \centering
    \includegraphics[width=\columnwidth]{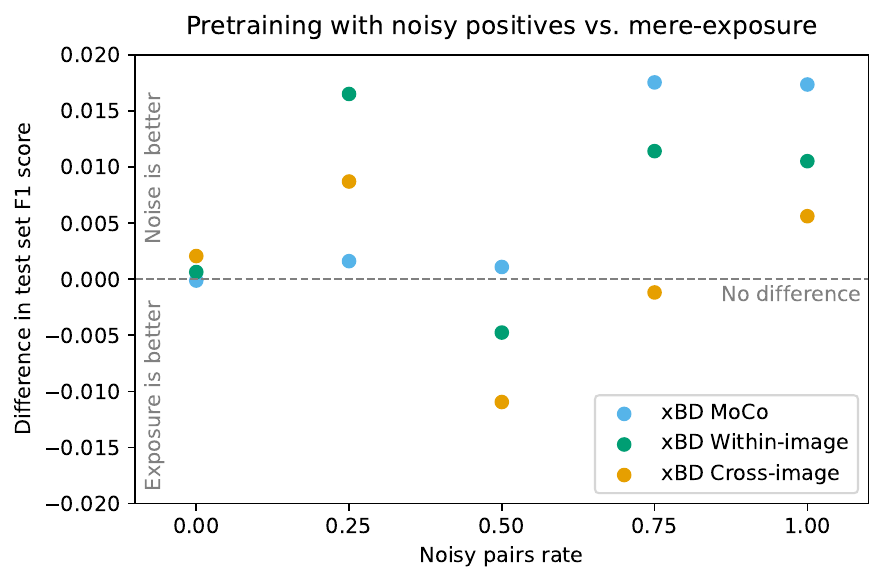}
    \caption{\textbf{xBD: Noisy positives pretraining tends to outperform mere exposure.} This shows that the small performance gains on xBD are not just attributable to exposure. Positive values mean that pretraining with noisy positives performs better on the downstream task than pretraining with mere exposure pairs, negative values mean the opposite.}
    \label{fig:xBD-exposure}
\end{figure}

\phantomsection
\label{sec:exp-xbd-exposure}
\textbf{xBD: More than mere exposure} The results on xBD are shown in \autoref{fig:xBD-exposure}. Noisy positives tend to perform better than the mere-exposure setting. We interpret this as an indication that the small performance improvements when introducing noisy positive pairs into xBD pretraining (see \autoref{sec:exp-xbd}) are not only caused by exposure to the building damage features, but also an additional effect.

%\footnotetext{The xBD dataset was released under the CC BY-NC-SA 4.0 license: \url{https://creativecommons.org/licenses/by-nc-sa/4.0/}.} %footnote belongs to noisy vs exposure fig

\begin{figure}
    \centering
    \includegraphics[width=\columnwidth]{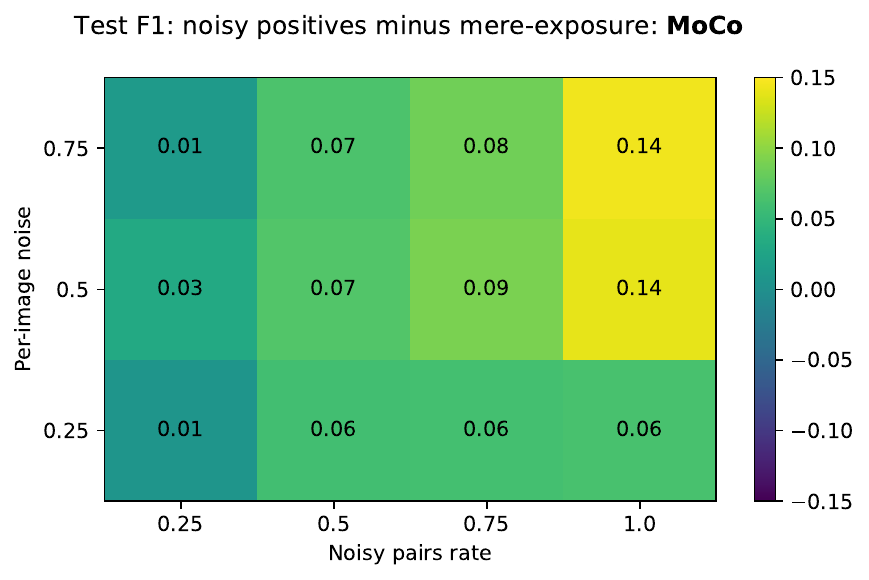}
    \caption{\label{fig:exp-voronoi-moco-exposure}\textbf{VTS: MoCo performs better with noisy positives than with mere exposure.} 
    Noisy positives outperform the mere-exposure setting, more so the more noisy pairs are included. On average, noisy positives pretraining achieves a $6.8\%$ higher test F1 score. This confirms that the positive effect of noise positives is not just based on an increased exposure to features that are relevant to the downstream task. Positive values mean that pretraining with noisy positives performs better on the downstream task than pretraining with mere exposure pairs, negative values mean the opposite.}
\end{figure}

\begin{figure}
    \centering
    \includegraphics[width=\columnwidth]{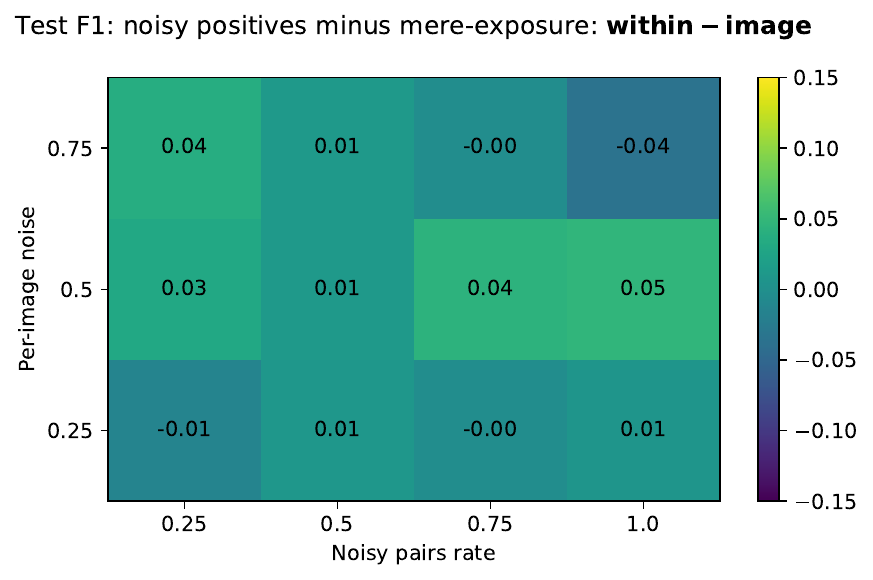}
    \includegraphics[width=\columnwidth]{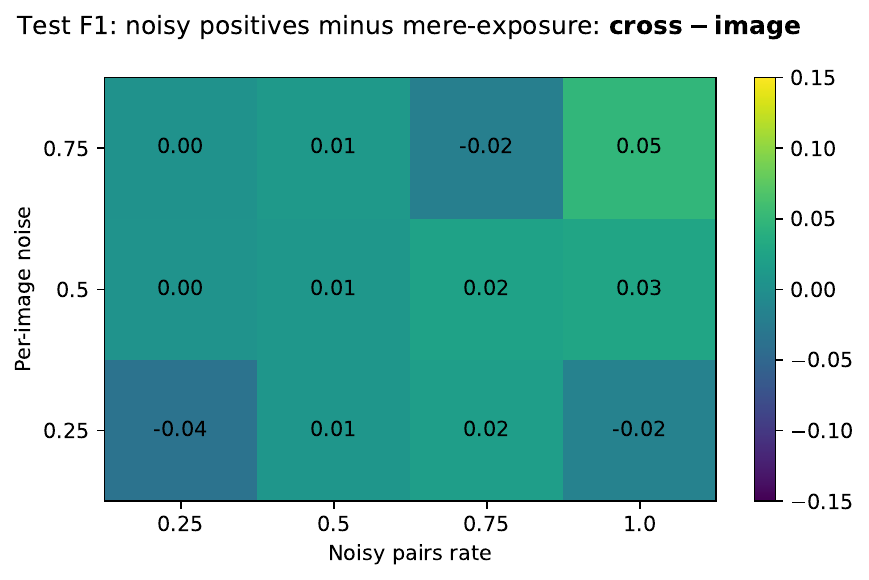}
    \caption{\label{fig:exp-voronoi-dense-exposure}\textbf{VTS: Dense losses on average perform slightly better when pretrained with noisy positives than with mere exposure.} Overall, the effect is much weaker than for pretraining with MoCo and several runs also performed better under mere-exposure than under noisy positives. On average, the within-image loss achieves a $1.1\%$ higher test F1 score with noisy positives pretraining, while the cross-image loss achieves $0.6\%$ on average.
    Positive values mean that pretraining with noisy positives performs better on the downstream task than pretraining with mere exposure pairs, negative values mean the opposite.}
\end{figure}

\phantomsection
\label{sec:exp-vts-exposure}
\textbf{VTS: MoCo benefits strongly from noise} For VTS, we define mere-exposure pairs analogously to xBD. \autoref{fig:exp-voronoi-moco-exposure} shows the performance difference between models pretrained with noisy pairs and those pretrained with mere exposure, using MoCo for pretraining on the VTS dataset. Noisy positives outperform mere-exposure with rising $r_{pairs}$, given high enough $r_{img}$. This aligns with the xBD results and shows that MoCo benefits from the noise introduced via the noisy pairs. The dense losses (see \autoref{fig:exp-voronoi-dense-exposure})
also perform slightly better on average using noisy pairs as compared to mere exposure pairs. The differences between both settings are much smaller for dense losses than for MoCo, however. For the within-image loss, we compute a mean F1 score difference of $\mu{=}1.1$\% and standard deviation $\sigma {=} 2.3$\%, where positive values indicate that pretraining with noisy pairs performs better. For the cross-image loss, we compute $\mu{=}0.6$\%, $\sigma{=}2.2$\%. In comparison, MoCo achieves a much larger$\mu{=}6.8$\%, $\sigma{=}4.1$\%.

\subsection{Discussion}

In this section, we discuss our experimental results related to the effect of including noisy positive pairs in contrastive pretraining, what we believe is responsible for the observed effect and how our results relate to existing research.

\textbf{Noisy positives improve performance via regularization} Overall, the increased performance of models pretrained with noisy positives is not only explained by an effect of mere exposure to relevant features, but by an additional effect that must be based on the noisy matches (see \autoref{sec:reguarlization}). We believe that this effect can be interpreted as a form of regularization akin to image augmentation. Pairing intact buildings with damaged buildings could be seen as a form of strong augmentation. In extreme cases of very different appearances between the two images, we approach augmentation regimes that resemble CutMix~\cite{yun_cutmix_2019}, which simply pastes part of one image onto another image. In extensive experiments, variants of this have already been shown to improve performance of self-supervised pretraining~\cite{ren_simple_2022,li_self-supervised_2020, lee_i-mix_2021}.

\textbf{Noisy vs. false positives} \autoref{fig:1} shows that \textit{false} positive pairs lead to declining performance, while \textit{noisy} positive pairs do not, and can even be beneficial, e.g. in \autoref{fig:exp-voronoi-moco}. We believe that the performance does not decline, because the model can still identify enough similarities between the unchanged parts of images to identify the matching images during contrastive learning. The positive effect is additionally caused by the regularization, as explained above. Furthermore, we see that when we move from partially noisy positives towards fully false positives, by setting the per-image noise $r_{img}=100\%$, the performance does usually drop, especially when all positive pairs are impacted by this noise, i.e. $r_{pairs}=100\%$. In this way, we can reconcile our results on noisy positive pairs with those on false positive pairs. An exception is the cross-image loss in \autoref{fig:exp-voronoi-dense}, which always receives a correct learning signal from its within-image loss component and improves performance as long as not all of the positive pairs are false positives. 

\textbf{Self-supervised dense losses work despite of noisy labels} Current self-supervised dense contrastive methods often use pseudo-labels~\cite{zhou_c3_2021,zhong_pixel_2021} or feature space clustering~\cite{hamilton_unsupervised_2022,zhang_looking_2021,wang_dense_2021} to compute dense contrastive pairs. If the robustness we saw for the supervised dense losses transfers to other dense losses, this might be an indication why these self-supervised learning methods work, even though they rely on noisy labels to create contrastive pairs.

\section{Conclusion}

We investigated the effect of noisy positive pairs in contrastive pretraining for semantic segmentation. For both datasets we investigated, and three different contrastive approaches, we were not able to create conditions under which noisy positive pairs are consistently detrimental. Instead, they are beneficial for MoCo and dense losses, although the effect is small on the real-world dataset. The only situation in which noisy positive pairs proved detrimental was the edge case in which noisy positive pairs become false positive pairs. 

We further found that the performance gains from pretraining with noisy positives are not solely caused by an increased exposure to features relevant to the downstream task. Instead, the improvement seems to be additionally attributable to an element of regularization, that is especially strong for MoCo. 

As a consequence of these results, practitioners do not need to take counter-measures against noisy positive pairs in contrastive pretraining. They should instead be encouraged to actively include such pairs in their pretraining, for example by purposefully sampling from areas and times with known changes of interest, instead of avoiding them.

\section*{Acknowledgements} 
The computations were enabled by the supercomputing resource Berzelius provided by National Supercomputer Centre at Linköping University and the Knut and Alice Wallenberg foundation. We thank Prof. Devis Tuia for his extensive comments on how to make this paper easier to understand.

\newpage
\printbibliography

\clearpage

%\iffalse
%\fi
%\pagestyle{empty}

\appendix

\subsection{Author contributions}

Following the CRedit taxonomy\footnote{\url{https://credit.niso.org/}}, the individual contributions of the authors are the following.
\begin{itemize}
\item \textbf{Sebastian Gerard}: Conceptualization, Methodology, Software, Formal analysis, Investigation, Writing - Original Draft, Visualization
\item \textbf{Josephine Sullivan}: Methodology, Resources, Writing - Review \& Editing, Supervision
\end{itemize}

\subsection{Questions \& answers}

\subsubsection{The performance on xBD is relatively bad, why should these results matter?}

We do not aim for a high absolute performance, but want to find out how the relative performance changes between models pretrained with different amounts of noisy pairs. To this end, we purposefully restricted the model's capabilities. The idea is to increase the influence of the noisy pairs, to find out how detrimental they can be in a worst case scenario. In high capacity models, parts of the network could memorize the noise, while other parts learn the correct signal. In finetuning, the subnetwork that learned the correct signal could then be used to achieve good performance. 

\subsubsection{False positive and negative pairs were shown to have a clearly negative impact for classification tasks. Why is it different for segmentation?}

An important difference between most of these experiments and ours is that our positive pairs are only partially falsely matched. Pairing a cat and a dog is different from pairing a cat with an image that's half cat, half dog. A variant of the latter is actually known as the CutMix method, shown to improve performance. When going to the extreme of completely falsely matched images, we also see a negative impact on performance, but for realistic cases the results are non-negative. For the partially wrongly matched images, the network is still able to find relevant similarities, providing a learning signal, which is not available when matching two completely different images.

\subsubsection{The noisy matches in xBD and VTS consist only of image features relevant to the downstream task. What if we use 'real' noise, that is not useful downstream?}

This was our first experimental setup for VTS. For MoCo, we saw that low amounts of per-image noise were slightly beneficial, while high amounts were detrimental. For the within-image loss, adding any noisy pairs led to improvements over having no noisy pairs. A per-image noise of 100\% was always detrimental. Apart from that, the results had some variance, but no clear patterns. For the cross-image loss, the variance was similarly large as for the within-image loss and also had no clear patterns, not even in the extreme values. %Due to our setup, the pretraining at this time was non-deterministic, so that even for 0\% per-dataset noise, the results showed some variance.

\subsubsection{Is it not unrealistic to only have noisy pairings between features that are relevant to the downstream task?}

No. Generally, this is the realistic setting. The model is pretrained on images that it should later be able to segment. The pixels in both images that we interpret as noisy matches during pretraining will have to be classified during downstream segmentation. Therefore, their features are all relevant to the downstream segmentation task.

%\subsubsection{What do these results mean for practitioners?}
%The results are an indication that when pretraining for semantic segmentation, noisy positive pairs do not have a negative impact. Practitioners do not need to take steps to filter them out of their pretraining dataset and might even consider actively adding them, to at least ensure that the network is exposed to the features that it is supposed to detect in the downstream task.

%\subsubsection{How do these results relate to the recent developments in contrastive methods?}
%Our results are in line with methods using pseudo-labels to compute pixel-wise pairings in a self-supervised way. These methods are showing promising results, despite the noise present in pseudo-labels, similar to our results showing that noisy positive pairs generally don't have a negative impact on performance. 

%\fi

%\subsubsection{How would methods without any negative pairs be impacted?}

\end{document}